# Detecting retinal disease with Accelerated reused Convolutional network (ArConvNet)

Amin Ahmadi Kasani[1], Hedieh Sajedi[2*]

[1]*Department of Mathematics, Statistics and Computer Science, College of Science, University of Tehran,* Tehran, Iran, aakasani@ut.ac.ir

[2]*Department of Mathematics, Statistics and Computer Science, College of Science, University of Tehran,* Tehran, Iran, hhsajedi@ut.ac.ir, Corresponding author

**Abstract** Convolutional neural networks are constantly being developed, some efforts improve accuracy, some increase speed, and some increase accessibility. Improving accessibility allows the use of neural networks in a wider range of tasks, including the detection of eye diseases. Early diagnosis of eye diseases and visiting an ophthalmologist can prevent many vision disorders. Because of the importance of this issue, various data sets have been collected from the cornea of the eye to facilitate the process of making neural network models. However, most of the methods introduced in the past are computationally complicated. In this study, we tried to increase the accessibility of deep neural network models. We did this from the most basic level, i.e. changing and improving the convolutional layers. By doing so, we created a new general model that use our new convolutional layer named ArConv layers. Due to the proper functioning of the new layer, the model has suitable complexity for use in mobile phones and perform the task of diagnosing the presence of disease with high accuracy. The final model introduced by us has only 1.3 million parameters and compared to the MobileNetV2 model, which has 2.2 million parameters, after training the model only on the RfMiD data set under the same conditions, results showed that it had better accuracy in the final evaluation on the RfMiD test set. 0.9328 Versus 0.9266.

## 1. INTRODUCTION

Vision is one of the most important senses in humans, according to the evolutionary characteristics of humans; vision is the largest system in brain and occupies 20–30% in of the cortex [1]. As a result, it has a great impact on all aspects of life, including health, the ability to learn and work, help to others and its absence has bad consequences and severely affects people's lives. Eye diseases can cause vision disorders and blindness, and people who live in vulnerable communities have less access to medical diagnosis facilities, which will make the problem bigger. Increasing medical diagnosis facilities costs a lot, so providing less expensive solutions will be very effective[2].

Eye diseases such as diabetic retinopathy, age-related macular degeneration and other similar diseases, are recognizable according to their visual symptoms. A morphologist can diagnose the disease by looking at the eyes directly or by analyzing the recorded images of the retina. In cases where there is no awareness of eye disease, people do not seek any treatment, which may cause blindness [3], [4], [5]. Due to the need for drugs or other treatments, people must visit a specialist doctor in any case; as a result, diagnosing the type of disease using automatic tools by ordinary people cannot be effective in reducing the risks of retinal disease. However, it can be effective if it speeds up the diagnosis process and reduces the cost of treatment and increase availability of doctors or specialists [6].

Due to the lack of ophthalmologists in communities that have many eye problems [7], the access of people to see a doctor is limited and the ophthalmologist has limited time to examine each patient. Recent studies [8] showed that in comparison, Artificial Intelligence (AI) models could have better accuracy and sensitivity versus what an ophthalmologist. In general, by making low-cost AI models with high accessibility we can increase the usability of the tool by volunteers or regular individuals and still have comparable performance for some diseases over ophthalmologist [9]. By doing so, identifying whether the eye has a disease can be easier, which leads to an earlier treatment by an ophthalmologist, doctors and prevent the progression of the disease. However, in this context, there have been limitations in the number of retinal image datasets available that have disease labels. In this research, we try to build an efficient deep neural network model of the Convolutional Neural Network (CNN) type with general use cases in mind. So that in addition to the small size, and the optimality of calculations, it has a suitable sensitivity to data sets with little data. We need a model that can extract general related information in the training phase, which results in having high accuracy in testing.

To achieve proper performance for deep neural network training, we need a large number of diverse images. Even if the number of training images is small, we cannot use the same training images with a slight change to evaluate their method, because that evaluation is not valid. In this study, we used the RfMiD dataset [10]; this dataset has 1920 training images, 640 validation images, and 640 images for the final test of the study. The images of this data set was captured by DSLR cameras with special lenses and then captured images where labeled by three ophthalmologists. This data set has 46 labels; the first label has assessed the risk of disease existence, which will be the most important label in this research in terms of the applicability of the solutions we can make with it. Other than that, this label has more samples, which increase reliability of experiment results. Other labels have represent different diseases, some of which have a limited presence in the training and testing datasets and even less than 10 samples. Other labels represent different diseases, some of which have a limited presence with less than 10 samples in the training and even fewer in the test dataset. Evaluating the model based on a small number of samples can lead to wrong judgments, which we should avoid. Therefore, we try to turn every diagnosis of the disease into a diagnosis of the presence of the disease.

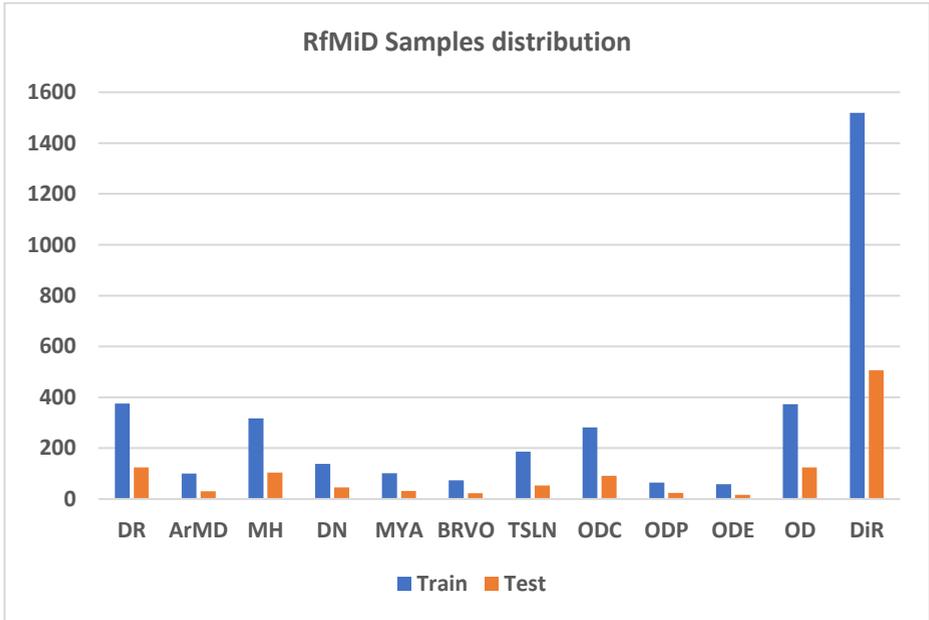

Fig. 1    We can see a large variation in the distribution of data. Note that the frequency of data for unlisted diseases, which are summarized in the others label, is much less than for listed diseases. In addition, each sample can have several diseases, so the sum of data frequency is not equal to the number of samples.

In the RfMiD dataset, the diesis risk category is larger than the occurrence of other categories. The images of this category may have unclear features that it is not clear the sample belongs to which diesis. Therefore, we use this category to evaluate the model because we think it is the most useful category. In other words, this category shows the characteristics of different diseases compared to healthy samples. Eye diseases have different appearance characteristics that can evaluate the performance of the solution in different ways. In the following, we will examine the characteristics of these diseases.The first disease under consideration is Diabetic Retinopathy (DR), often undetectable by the person themselves until vision is seriously damaged [11]. As a result, timely diagnosis of the disease is very important. Although blood tests and information about the occurrence of relatives' diseases can help in diagnosing the disease. Deep CNNs can recognize the appearance of the DR disease features located in the eye samples [12]. The second disease is age-related macular degeneration (ArMD), which is very common even in developed countries. Increasing age increases the risk of infection. Seeing tiny yellow or white sub retinal deposits is often one of the first signs of ArMD [13]. Other evidences of the occurrence of this disease is not obtainable by only eye examination. Including the patient's history, face-to-face tests including questions and answers, which can help to increase the accuracy of disease diagnosis, but they are not present in the eye disease data set.

The next abnormality named Media Haze (MH), which causes clouding of vision and is identifiable with cloudiness in eye images [14]. The diseases that cause this abnormality include cornea, aqueous humor, lens, and vitreous humor. As a result, identifying this abnormality does not solve the entire problem because each problem have different treatments. The next abnormality is Druses (DN), which are small yellow deposits of protein and lipids (fat) that form under the retina [15]. These symptoms are similar to ArMD and can be early signs of ArMD occurring in the eye, so a patient may have druses but not have ArMD. The next abnormality is Myopia (MYA), that often seen due to choroidal thickness changes [16]. Myopia can have a huge impact on people's quality of life. There are other symptoms include blurriness, fatigue, and blinking to see clearly. Branch retinal vein occlusion (BRVO) abnormality has signs of blockage of one or more branches of the central retinal vein, small bleeding, flame-shaped bleeding, soft and hard secretions, retinal edema, and dilated tortuous veins. Other symptoms of this disease are blurred vision and bleeding spots in the eye [17].

The main symptom of tessellation anomaly (TSLN) is the observation of large vessels in the choroid of the eye [18]. Although a small amount of this abnormality occurs with age, but mostly MYA is the cause. Optic disc cupping (ODC) also known as optic nerve head (ONH) is a serious problem and need immediate treatment. Increased intraocular fluid pressure on optic disc, which is the area where the optic nerve enters the retina [19], is the main cause of this abnormality. Increased pressure damages eye nerve cells and causes a blind spot in the eye. Optic disc pallor (ODP) refers to the abnormally pale appearance of the optic disc. The optic disc is usually pink in color with a central yellow depression, but in ODP, it is yellow. This disease can cause blind spots and even blindness [20]. Optic disc edema (ODE) is detectable by swelling of the optic disc and can be a symptom of various underlying conditions of other diseases. The sign of this abnormality in the fundus images is the absence of the optic disc [21].

Note that only one dataset alone cannot represent the proper performance of the introduced model. According to the information and features that we know about eye disease, we use the ODir-2019 dataset to evaluate the model. This dataset contains fundus images from each one of the eyes of 5,000 patients (two samples from each patient). In this study, we only use training samples including 3500 records. The ODir-2019 include eight category and the eight categories are assigned to the labels corresponding to each eye. These categories include normal, retinopathy, glaucoma, cataract, age-related, hypertensive, myopia, and others. Since we did not have access to the test images of this dataset, we divided the training data into two parts. We randomly selected 700 images for testing and used the other 6300 images for data augmentation and then training the models.

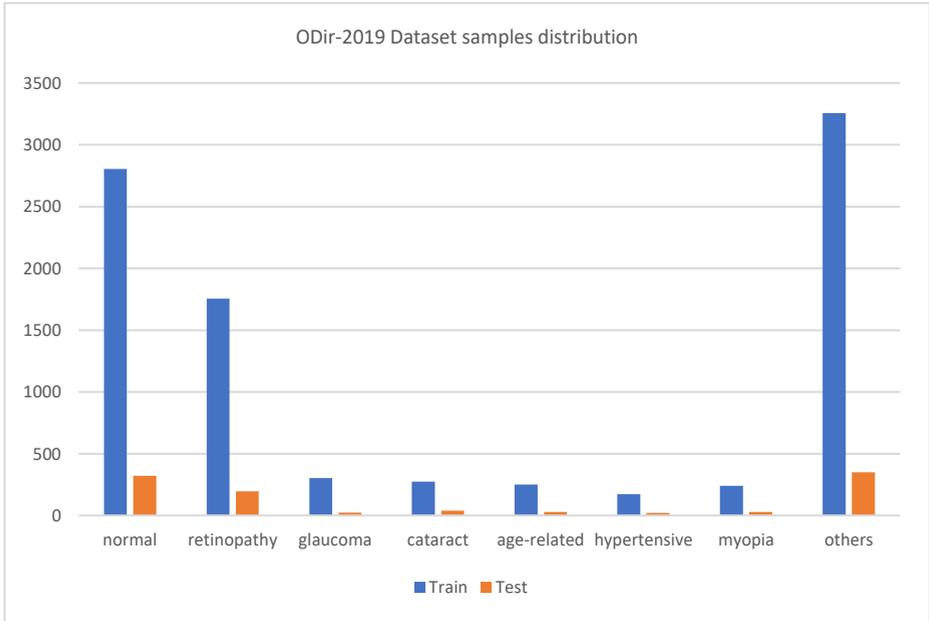

Fig. 2   This diagram shows the distribution of samples in different categories in ODir-2019. It can be seen that the others category has a large number of examples in this dataset, which is consistent with our impression in the RfMiD dataset.

The quality of training and test data segmentation is not very important, because in this study, we evaluate all models against this segmentation in the same situation in term of train and test parameters and number of samples. As a result, we should only compare the results obtained from the performance of the models in this study. To simplify comparisons, we converted the problem to binary classification. One class includes images of healthy samples and the other class includes images with abnormalities.

## 1. RELATED WORKS

In this study, we have addressed two areas, the first area is the construction of an optimal deep neural network model with high efficiency, and the second area is the diagnosis of eye diseases to evaluate this new model. According to the structure of the model, with an initial look at this concept by experts familiar with deep neural networks and image processing, it seems that our model will perform poorly. However we choose this dataset to show the opposite because of its rounded features where the spatial connections between the pixels of the images are very important.

1. Studies in the construction of general neural networks with efficient processing cost

In this research, many studies inspired us, from each of the mentioned models; we use a lot of proven information to build out optimal model. CNN models, including AlexNet, were initially made of simple 2D convolutional layers. A simple convolutional 2D layer tries to create filters in a three-dimensional space, with two spatial dimensions (width and height) and to learn the channel dimension. Therefore, a single convolution kernel has the task of simultaneously mapping inter-channel correlations and spatial correlations. Models such as ResNet [22] still used these Convolutional layers but with smaller repercussions with the help of residual connections and bottleneck design. Next solution was grouping convolution layers, early attempts happened in Inception V1, which used different parallel layers with different kernels in each block and concatenate them in the end. However, these solutions still required a lot of calculations and parameters for proper performance. There were ideas to make parallel layers in each block more efficient, including in Asymmetric models [23]. Making parallel layers smaller, instead of a 5*5 convolution and other expensive layers was one of these solutions. They separated spatial dimension the additional parallel layers where working on by adding 1D kernels that are a specific spatial dimension such as height or width. This process continued in the InceptionV2 and InceptionV3 module [24], by creating parallel layers but with smaller processing cost.

On the other hand, there have always been efforts to make the models smaller and better optimized because reducing the computational cost could make deep neural network models available to more people, and it was possible to run them on more hardware. Among the attempts to summarize, the famous and large pre-trained models [25] . ResNext [26] introduced group in Convolutional layers, it is similar to parallel layers in Inception model but all parallel layers are similar. This strategy results in speed up in computation because we can compute all the parallel layers at the same time. as well as enrichment of extracted features and reduce computational cost. [27] Introduced the idea of Depthwise Convolutions by using 2D Convolutional layers followed by 1*1-convolution layer. Later the idea of using group in convolutional layers went to next level and by setting the number of groups equal to input channels. This improvement showed its potential with the name of Separable Depthwise Convolutions in Xception [28] and became more prominent in many other studies. Separable Depthwise Convolutions made the number of parameters of the models more concise by completely separating computation of spatial information in (height, width) and channels. We could deepen the calculation blocks without having a negative charge of high number of parameters and complexity, which comes with it. The key of improvement is that we could make our model deep with richer feature extraction. [29] Has already showed that resnet101 it has similar functionality to resnet1000, although it is deeper and has more parameters [29].

SqueezeNet [30], used the idea of using bottleneck design that first ResNet models introduced, in extreme level to reduce the size of model blocks as much as possible and still achieve good results. On one hand, it reduced the expensive part of using a 2D convolutional layer for extracting spatial features, and on the other hand, this summarization reduced the overhead and made the information in the layers more general with very limited parameters. Therefore, DenseNet [31] used these and by introducing dense residual connections on top of that, they reduced the number of parameters in each of the blocks even more, which resulted in better performance and accuracy than ResNet. In terms of the optimality and cost-effectiveness of these ideas, they matured a little in MobileNetV2 [32], by creating an inverted res-block and optimizing the number of parameters by using a separable Depthwise convolution design and squeeze and expansion layers in each block. This idea, on the one hand, reduced the number of model parameters when collecting the output of each residual connection and reduced the memory load of the model. Because we have to keep each of the input of each block, in the memory and it remains unused until the block calculations are complete and its output is collected. On the other hand, they increased the optimality more and more because by connecting the input and output of big layers with high number of parameters to layers with low number of channels and mostly 1*1 convolutional layers. The EfficientNet[29] model took past ideas one step further and proposed the idea of making the convolutional channels wider and deepening the blocks and showed that with the ideas of MobileNetV2 by making it bigger, the best convolutional model can be made up to that time.

ShuffleNet [33] introduced ideas such as shuffle channels to improve the computational cost and efficiency. In addition, channel segmentation [34] has been able to reduce the computational cost like groups in two-dimensional convolutions. Image processing models based on transformers including Swin [35], [36] and Vit [37] are not convolutional models, but they have shown great performance at the time of introduction. These models are relatively costly in terms of the number of parameters and calculations. However, the use the ideas such as embedding two-dimensional inputs by dividing the input images into tiles. This idea can reduce the computational cost of layers and Summarize the blocks of convolutional models strongly. Among other useful ideas in these models is reducing the number of Normalization and Activation layers, which can increase the performance of convolutional models, among other ideas is the use of larger kernels and increasing the size of the expansion blocks at greater depths. ConvNext [38], [39] use these ideas in a CNN model and have shown very good performance.

## 2. Related works in the diagnosis of eye diseases

In the past, many studies tried to diagnose eye diseases with the help of image processing and convolutional neural networks. Many efforts focused on diabetic retinopathy due to the importance of this type of disease. Because this is a 4-stage disease and in the initial stage, it is possible to treat it before damage to the eye [40]. In the study [41], investigated genetic based ideas for disease diagnosis on different eye data sets. Among them is the RfMiD dataset, they have looked at this dataset as a multi-label problem, using the Resnet18 model; they have reached the result of 0.9362 ROC curve accuracy.

In the [42] study, five popular models were used to classify fundus images into healthy and diseased categories, including Inception V3, DenseNet-121 and EfficientNet, the ROC curve accuracy of these models was between 0.9477 and 0.9587. Finally, by assembling the models, they have achieved an ROC curve accuracy of 0.9613. In the field of classification for identifying diseases, the ensemble of models has achieved an average ROC curve accuracy of 0.9295.

In the [43] study, an attempt has been made to provide a solution for the diagnosis of age-related macular degeneration, so that it can be used with the help of smartphones. The approach used in this method includes the use of the deep neural network introduced in [44] to detect the macula, and then the Radon transform is used for preprocessing the images and classification of the resulting output with the help of SVM. The classification results of this approach for detecting ArMD disease in the RfMiD dataset have reached 94.3%, and it the computation duration is 15ms on the Galaxy S9 smartphone.

The study [45] has evaluated various popular models to identify fundus diseases from images. Meanwhile, they have obtained the best result with the help of a double ensemble setup of the EfficientNetB3 model. They used the (International Competition on Ocular Disease Intelligent Recognition) Odir-2019 dataset, this dataset has 8 categories including the healthy category, DR, and ArMD category, and finally the introduced model has 90% accuracy on the validation dataset and 67% been achieved on the test data set. [46] Uses multiple datasets including ODIR-2019. They create new training images using a GAN algorithm and then perform the classification work by CNN neural networks including ResNet-18. They achieved 85% accuracy in detecting AMD fundus images.

.

## 2. METHOD

One of the most important parts of image processing is the pre-processing step. Given the purpose of building an optimized and low-cost processing method, we keep the process low-cost in all the pre-processing steps. Given that a regular camera and a special lens record Fundus images, much of the input image is the worthless space of the lens. However, the brighter feature of the images can help us divide the input image into two general sections by selecting an Adaptive Threshold. Next, we find the geometrical boundaries of the eye in the binary image and finally cut the original image according to the boundaries found. Due to the noise in some images, choosing the right threshold for binary images helps reduce the error in cutting.

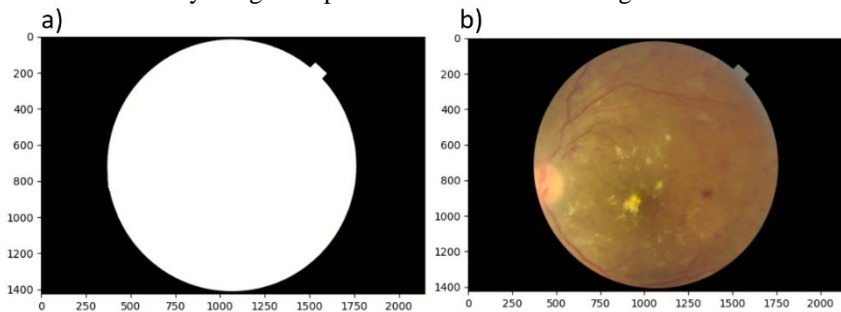

Fig. 3    The (b) image shows a raw input that we are using in this study, captured with a DSLR camera and special lens. The image on the (a) shows a binary mask created to cut out the main part of the image. An accurate binary mask has the ability to remove the noise around the fundus and we accurately easily and can cut the main part of the image with bright areas of the mask.

In the Introduction, we explained that there are a relatively small number of images to achieve excellent results in the dataset, especially for some diseases, whose number of training samples is less than 75 samples and they are not unique, because multiple diseases can be present in one image at the same time. In the Data Augmentation process, we made the images needed to improve the performance of our deep neural network models, in addition to making changes including rotation, change of light intensity, contrast change, color change, and zoom. We increase the number of samples with this method only once and use the final dataset for all test models. We should mention that we also did this for the ODir-2019 and RfMiD datasets separately.

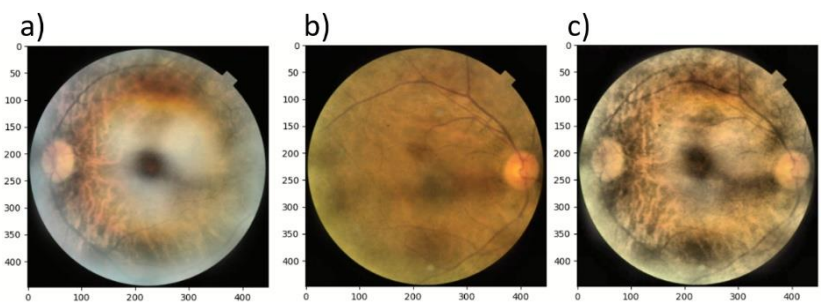

Fig. 4    Sections a, b show the input images of the integration algorithm. Section c shows the result of this integration, whose labels will include the active labels of both input images.

To improve the results, we decided to integrate images together to add more diseases into the training images. In other cases, we created more new and general patient specimens by combining healthy samples and samples including disease. We tested two ways in implementing this solution: In the first method, we created a collaborative environment with two models. We gave the first model the task of combining images, and the second model the task of classifying patient and healthy images of Fundus. For training the models, we updated the weights so that the first model of images could create clear information from the diseases of its two input images so that the second model could correctly detect. The result may be able to provide ophthalmologists with clearer information on the symptoms of the disease, which we can continue in another study. In the second method, we integrated two distinct input images with the average of the corresponding pixels, which results in a new image with diseases presented at each of the inputs. The second method is much easier than the first method; the disadvantages of this method are reinforcing the features that we can ignore and is present in two input images. The second problem with this method is that it can eliminate some information after merging the two input.

The use of data augmentation at the input level of this method of combining samples can lead to the creation of images with more complete patterns and provide more states than the direct use of this method for the classification model. In short, using inputs with changing angles, color and different contrasts as the input of this method is better than using the initial input and data augmentation on the outputs of this method. After comparing the results of the two types of creating new artificially ill samples by employing evaluation in different models, we saw that in some cases the marks were more detectable for the human eye, but did not have a significant impact on the diagnosis of the disease by the classification model. Therefore, one of the applications of this model can be to show the appearance patterns related to diseases more than what can be seen directly. Given that in this study, our goal is to achieve maximum accuracy with minimal use of resources, we continued to integrate images using the second method.

We know that neural networks with low depth have a high potential for overfitting and not extracting the connection between the input data and the desired labels. In addition, according to the objectives of this study, we do not want to use very large neural networks, so we try to create a new deep neural network with the right number of parameters to run on mobile phones and compare it with the most famous light neural networks, such as MobileNetV2, MobileNetV3, and EfficientNetB0 to compare. The advantage of using deep neural networks with a low number of parameters is the maximizing extraction of the desired information from the input data.

Based on our research and previous related studies, we explained how the use of methods such as Inverted Residual Block and the use of Separable Depthwise Convolutions instead of conventional convolutions in the past helped to reduce the number of parameters. In addition, using less activation and normalization layers can help improve the final accuracy of the model. In addition, controlling the size of Expansion Layers can have a great impact on the complexity and efficiency of our neural network model. At this stage, we are trying to build an optimal model based on the knowledge and information obtained.

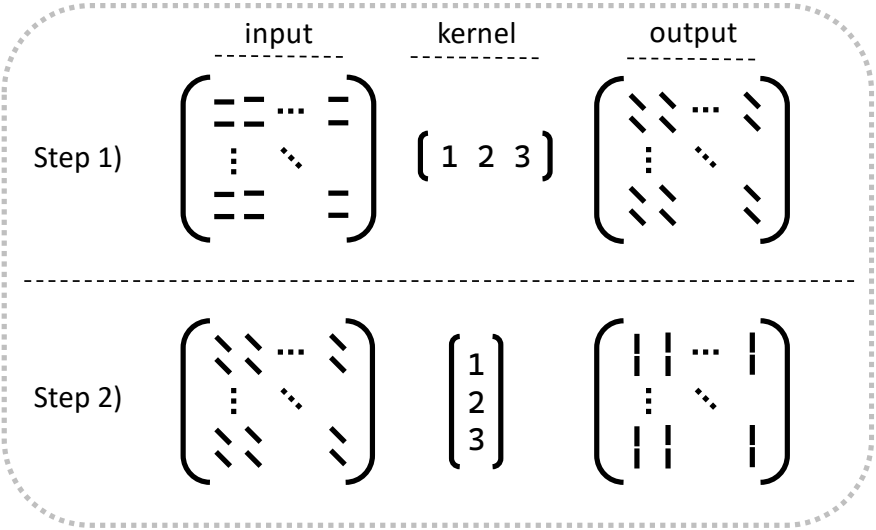

Fig. 5  In (step 1) we apply a 1D Depthwise kernel to the input and extract features inside rows. In (step 2) we get the output of the step 1 and apply transpose of the same 1D Depthwise kernel to achieve final output. The above figure shows the order of kernels but in a big model, the order does not change the results; in this block.

Now we will introduce the Accelerated reuse Convolutional Layer (ArConv Layer). We know that a Separable Depthwise Convolution layer is a convolution layer whose number of its group is equal to the number of its filters and the number of filters is equal to the number of input channels. The size of the kernels of this layer is two dimensions or more, and the kernel size is 3 or more. If we consider the size of the kernel of this layer to be one-dimensional like (1*3 or 3*1), the connection and information between the columns of the input data will be completely isolated from each other. This work is similar to applying a 1D convolutional layer as a time series on the input data so that the number of groups is equal to the number of filters and the number of filters equal to the number of input layer channels. Now, to solve the problem of isolation of the input lines from each other, we use the idea of repeated use of a layer and transposition of column and row of input. We create a two-dimensional separable Depthwise convolution layer, apply it to the input data, and then transpose the rows and columns of the output of this layer with each other. In the next step, we apply the same layer on the data and this time we extract the relationships between the rows of the input data. Finally, we change the transpose of the row and column output of the previous layer once more and get the final output. According to the size of the kernel of this layer, we can save more than 66% of the number of parameters which results in lower number of memory usage and model being smaller.

a)

$$input: \begin{bmatrix} 1 & 2 & 1 & 1 & 2 \\ 1 & 2 & 2 & 1 & 2 \\ 0 & 1 & 0 & 1 & 1 \\ 0 & 0 & 2 & 0 & 0 \\ 1 & 2 & 1 & 2 & 2 \end{bmatrix} kernel: \begin{bmatrix} 0.3225 & 0.4578 & 0.6938 \\ 0.9707 & 0.7073 & 0.6712 \\ 0.7724 & 0.6559 & 0.8607 \end{bmatrix} output: \begin{bmatrix} 3.7516 & 5.3113 & 5.6466 & 5.8106 & 3.6956 \\ 5.1721 & 7.4303 & 7.8942 & 7.8333 & 5.4103 \\ 3.1715 & 5.0698 & 6.3636 & 6.9881 & 3.5511 \\ 1.8753 & 4.5294 & 5.5385 & 5.0256 & 3.8201 \\ 2.6487 & 4.6016 & 5.3031 & 5.7488 & 2.7572 \end{bmatrix}$$

b)

$$input: \begin{bmatrix} 1 & 2 & 1 & 1 & 2 \\ 1 & 2 & 2 & 1 & 2 \\ 0 & 1 & 0 & 1 & 1 \\ 0 & 0 & 2 & 0 & 0 \\ 1 & 2 & 1 & 2 & 2 \end{bmatrix} kernel: \begin{bmatrix} 0.7299 \\ 0.8761 \\ 0.8813 \end{bmatrix} output\ of\ first\ apply: \begin{bmatrix} 1.6060 & 3.2121 & 2.3360 & 1.6060 & 3.2121 \\ 1.7574 & 4.2448 & 2.6335 & 2.4873 & 4.2448 \\ 0.8813 & 2.6387 & 3.2225 & 1.7574 & 2.6387 \\ 0.7299 & 2.3412 & 2.4822 & 2.3412 & 2.3412 \\ 0.8761 & 1.7522 & 2.6387 & 1.7522 & 1.7522 \end{bmatrix}$$

$$final\ output: \begin{bmatrix} 3.7518 & 5.9349 & 6.0499 & 5.8106 & 4.2297 \\ 4.6382 & 7.1902 & 7.8640 & 7.5988 & 5.9111 \\ 2.6983 & 5.4409 & 6.4317 & 6.3060 & 3.8607 \\ 2.3485 & 4.5064 & 5.9478 & 5.9478 & 4.1145 \\ 2.0466 & 4.2335 & 5.1352 & 5.1398 & 3.0794 \end{bmatrix}, mean\ absolute\ error:\ 0.3678$$

Fig. 6     (a) Shows the convolution of a two-dimensional kernel on the input. (b) We attempted to find a one-dimensional kernel that replicate original kernel behavior, so that the results of the ArConv application are as close as possible to the initial output.

However, the main question that arises is what is the strength of these two layers relative to each other? To answer this question, we implemented two convolutional weight approximation algorithms corresponding to the dimension of kernels. The task of these algorithms is to estimate the convolution weights that lead to the output from the given input. Therefore, if we apply a 2D random kernel as a convolution on the desired 2D input and get the corresponding output. Then we should be able to find the kernel in a one-dimensional way, which we did. The one-dimensional weight applied in Figure 6 is an example of the performance of this algorithm. In an experiment, we created a fixed random input corresponding to each input matrix size, then created 25 random two-dimensional kernels, and calculated its convolution output. Then we tried to find one-dimensional kernels that, when applied as ArConv on fixed input, give the same output as the two-dimensional kernel.

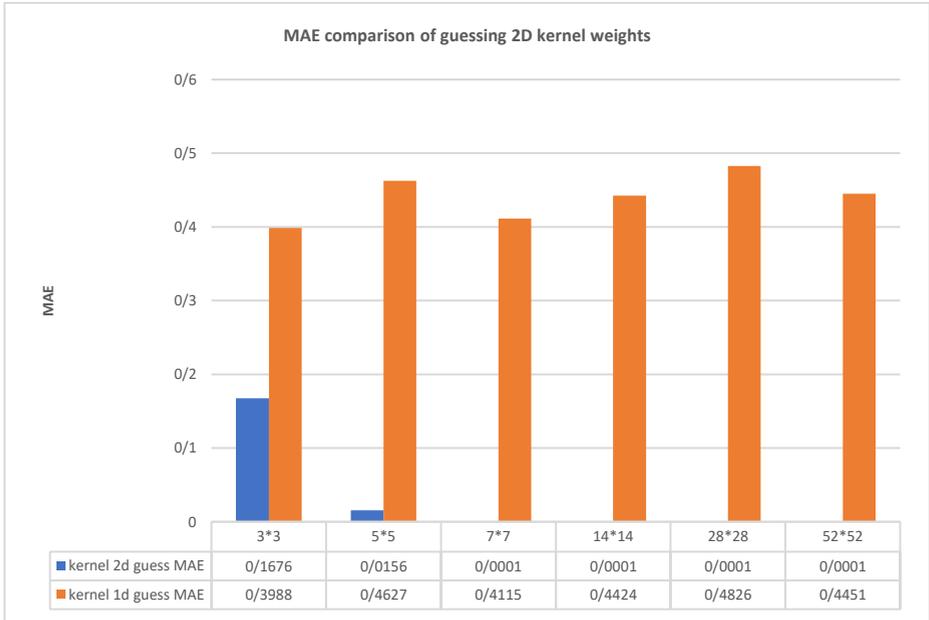

Fig. 7   The above graph shows the comparison between the performances of convolutional weight estimation algorithms. We use Two-dimensional kernels in convolution and one-dimensional kernels in ArConv to calculate the output error. From the input size of 7x7 and larger, the MAE obtained from the estimated two-dimensional kernel is about 0.0001.

By analyzing the Figure 7, it we can see that the power of the ArConv layer is lower than that of a convolutional layer, but the amount of error in reproducing the behavior of a 2D convolutional layer is not so great. This can cause limitations in the models built with this layer and prevent overfitting the kernel in training phase of the model. Therefore, in the following, we will try to show that we can ignore this error and we will further examine the advantages of this layer. Another question is that despite the 66% reduction in the number of parameters, does applying convolution twice in ArConv make the computation slower? In the tests performed by applying 100,000 normal convolutions and ArConv convolutions on a fixed input, the computation time for normal convolution was 0.30 seconds, and for ArConv convolution was 0.51 seconds. This is because the computation of a 2D convolution can be computed in parallel, while ArConv computation does not currently have this speedup, but the lower memory bandwidth consumption may have contributed to ArConv's speed. So one has to choose whether this trade-off is in favor of the problem at hand or not. In the next experiment, to make the power of ArConv and Conv2d more clearly, we generate the reference kernel using ArConv and try to guess it by another ArConv and Conv2d and compare the error in Figure 8. The results show that although the representation power of different modes is higher in Conv2d than ArConv. However, another ArConv layer has higher accuracy than Conv2d has done the calculation of weights that have the same behavior with the original ArConv layer. The reason for this is that the entered input can affect ArConv kernel behavior and is somewhat dynamic.

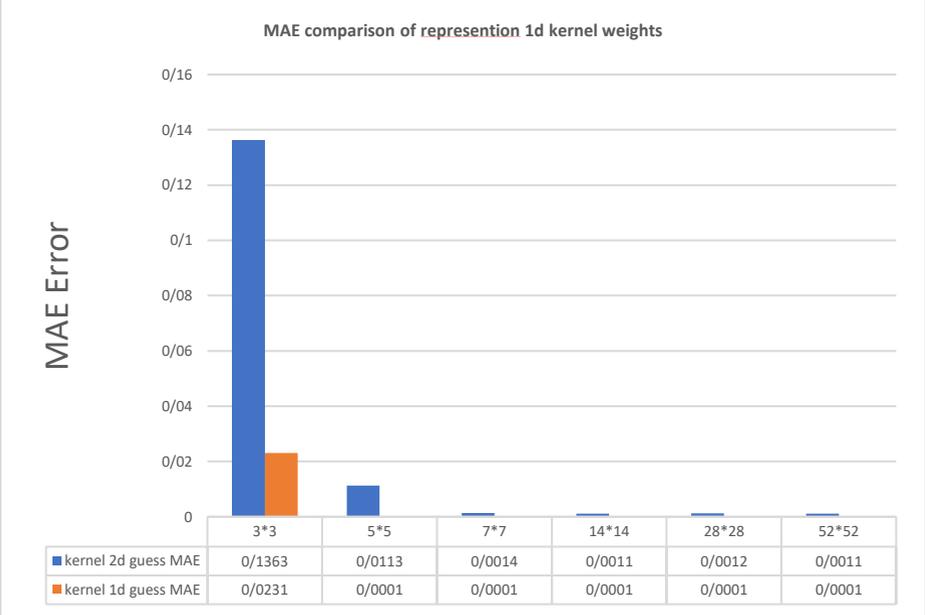

Fig. 8    This data shows that with the same settings, we can replicate the behavior of the original ArConv layer more accurately by an ArConv than by a Conv2D that has more parameters.

Now we introduce the ArConvNet computing block, this block starts with a dense layer that expand channels space. In the next step, if it is necessary to reduce the size of the data, we use a Separable Depthwise convolution with Strides=2. Otherwise. We use an ArConv Layer and finally, by using another dense layer, we summarize the information in this module and take it to the output. Due to the use of ArConv Layer, the use of large expansion values helps to increase the accuracy of the model, but the total number of parameters of the block is highly smaller than blocks in similar models.

Considering that, the computational cost of each Convolution layer has direct relation to the size of the kernel, the number of filters, and the size of the input data. The current neural networks spend a lot of money on the initial layers that have a large size, and after a few residual blocks, they try to Increase the input data size from 224 * 224 pixels to 56 * 56 pixels. In neural networks based on transformers, we saw that using the concept of the window. In this method, each part of the image is encoded and converted into primary data for processing, and after using the concept of the window, the problem of the computational cost of image data for processing in neural networks based on fixed the transformer and this technique did not cause poor performance of the model.

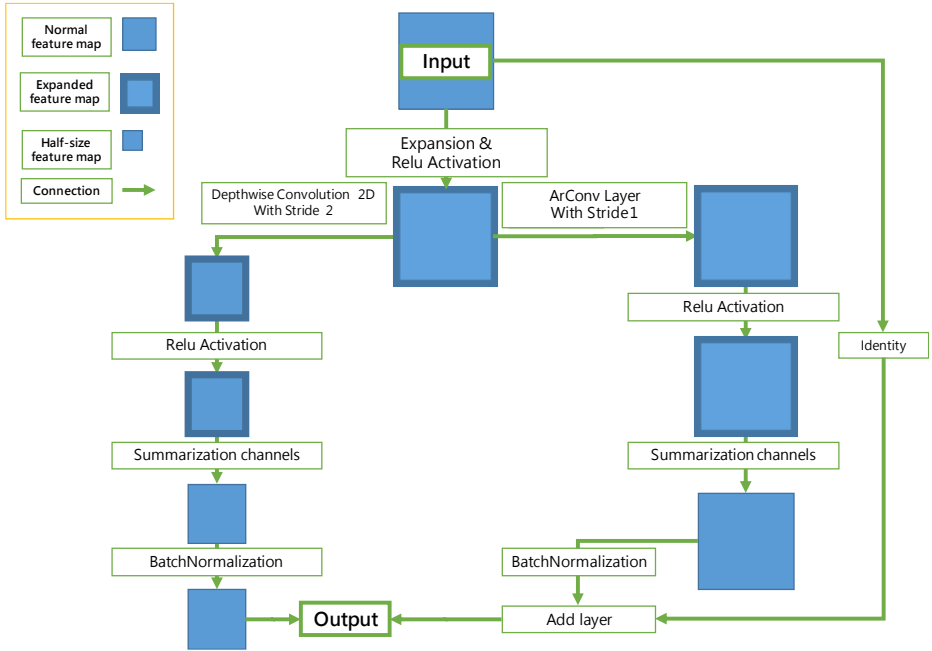

Fig. 9  The figure above shows a view of how ArConv Block is constructed. The input goes through a Relu Activation layer, and then if Stride is equal to two, it enters the left path. A depthwiseConvolution2D layer is applied to it, and then it passes through a Relu Activation layer, after that its number of channels is summed. Finally, after a Batch Normalization application on the data, we use the output as the input of the next block. If Stride is equal to one, the right path is used and we apply ArConv Layer instead of depthwiseConvolution2D. This path creates a residual block that we use between down sampling, so we add the input of the block to the output of the last layer.

Using the knowledge obtained from the Swin and similar models, we try to summarize the input layer in and down sample it from 224 * 224 pixels to 56 * 56 pixels with a simple two-dimensional convolutional layer. We use a Strides 4 and 4*4 kernel to compress image and do the sampling for encoding. However, the choice of the number of filters can have a big impact on the quality of the encoded windows. By choosing the right number of filters in our model, reducing the complexity in the initial layers of the model to reduce the size of the input data in one-step has greatly reduced the cost of calculations and has no significant effect on the extracted information.

Now we can see how we spend the amount of processing costs savings in the initial layers and blocks of the model for larger expansions in the inverted bottleneck residual blocks. In our deep neural network, we expect general information to be extracted in the initial blocks, that is why the expansion channels don't need to be big, as the depth increases, the expansion size also increases, the values used are listed in Table X.

Table.1.  The following table shows the construction information of ArConvNet model by ArConv Blocks.

| Block numbers | Input size | Output size | Expansion size | Type |
|---|---|---|---|---|
| 0 | 224*224*3 | 56*56*24 | - | Conv2D |
| 1-3 | 56*56*24 | 56*56*24 | 4 | ArConvBlock |
| 4-6 | 28*28*24 | 28*28*32 | 8 | ArConvBlock |
| 7-14 | 14*14*32 | 14*14*48 | 12 | ArConvBlock |
| 15-17 | 7*7*48 | 7*7*96 | 16 | ArConvBlock |
| 18 | 7*7*96 | 7*7*1536 | 16 | Dense |
| 19 | 7*7*1536 | 1536 | - | Global average pooling |
| 20 | 1536 | Classes | - | Dense Output |

The final model that we built without considering the two final layers has 1,316,376 parameters, and it has shown excellent performance considering the number of parameters and the depth of the model. We can change the final two layers of the model according to the type of problem and size of output. For example, as showed in Fig.10 in the binary classification, we use a global average-pooling layer, followed by a fully connected layer of size 2. Also, Adam optimizer with learning rate of 0.0001 and Cross entropy error that expect logit input were used for training, although the difference between Cross entropy error and mean squared error(MSE) on the accuracy of the results in the test data set obtained in binary classification was very small. To train the model in multiclass, we used the output layer equal to the number of classes and MSE error.

## 3. RESULT AND ANALYSIS

Our goal in this research is to introduce a general model to perform various tasks for which deep convolutional neural networks may be used. The selected field is due to the properties of the inputs that may be among the limitations or the place where this model shows poor performance according to the limitation our model can have. Because in the feature extraction process, we completely separated the dimensions of the image, both in the spatial space and in the channels space. For this reason, in this section, our priority is to compare the introduced model with well-known general models in binary classification setup.

In this section, we have trained all the models on the same data set with the same number of input images and the same sizes, and we have reported the best performance. To evaluate the models, we use weighted accuracy and weighted precession metrics (in the following, we will not mention weighted, the meaning is the same). In comparing the depth of models, we consider the number of Convolution or Dense layers with learnable parameters. We evaluated all of the models in terms of speed by running the experiment on a computer with NVIDIA RTX3080 GPU, Intel 12400f CPU, 64GB DDR4 memory. The ArConvNet model introduced by this research has the depth of 64 trainable layers, and has 1,319,488 parameters and its packaged size is 16,037KB. The calculation of a batch of 32 input images with the size of 224*224*3 in ArConvNet model finished in 16ms. This model has achieved 93.28% accuracy and 93% precision in the RfMiD test set. The next model is MobileNetV3Large, this model has 2,998,272 parameters in the binary classification setup, the model depth is 48, and its packaged size is 36,170KB. The calculation of a batch of 32 input images with the size of 224*224*3 in this model is finished in 15ms. The trained model has an accuracy of 92.19% and a precision of 92%.

Another model is MobileNetV2, this model has 2,260,544 parameters in binary classification setup, model depth is 36, and packaged model size is 27,073KB. MobileNetV2 finished the calculation of a batch of 32 images with the size of 224*224*3 is finished in 14ms. In the binary classification of disease risk, this model has achieved 92.66% accuracy and 93% precision. The fourth model is EfficientNetB0, this model has 4,052,131 parameters in the binary classification setup and is larger than the previous models, the depth of the model is 82 and the size of the packaged model is 48,306KB. This model finished the calculation of a BATCH of 32 images of 224*224*3 in 17ms. In the evaluation of the RfMiD test set, achieved an accuracy of 91.87% and a precision of 92%. The last model is the EfficientNetB3 model, this model is the larger model of its family, the EfficientNetB0 model. The number of parameters of the model is 10,786,607, the size of the packaged model is 127,538KB and the depth of the model is 131. Calculation of a batch of 32 images of the 224*224*3 size by this model in 28ms and is much slower than other models. In the RfMiD test, the developed model achieved 92.03% accuracy and 92% precision. The summary of the above tests is showed in Figure 10.

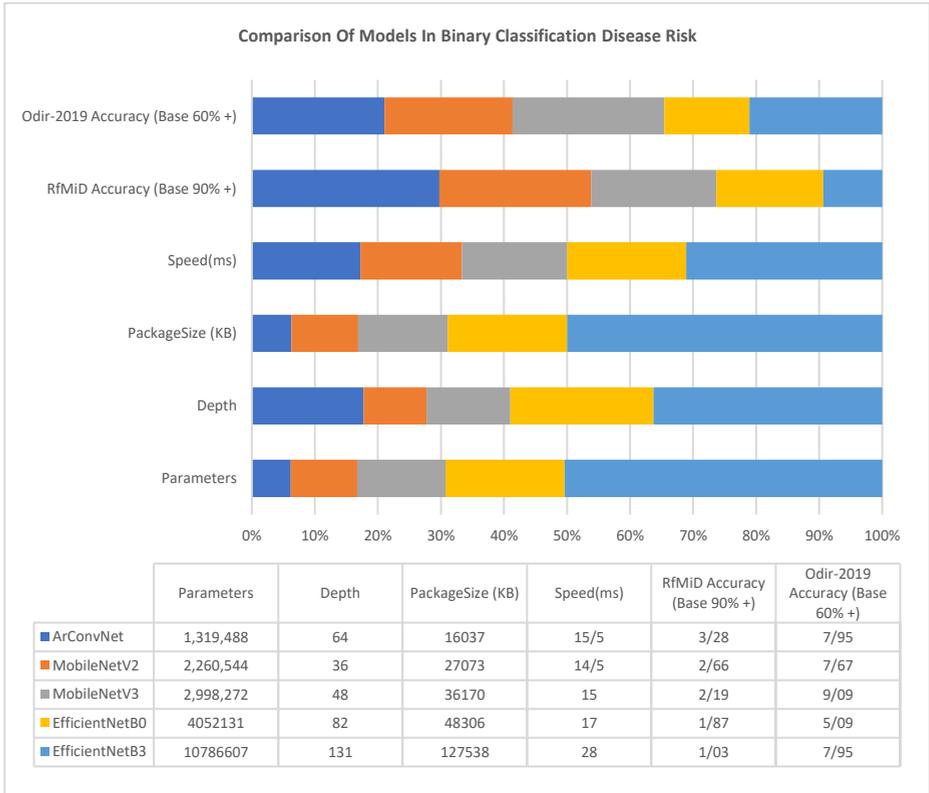

Fig. 10  The above figure shows the comparison of measurement and evaluation parameters of different models compared to the ArConvNet model. By looking at this diagram, we can clearly understand the relationship between the accuracy and depth of the model, as well as the number of parameters, speed, and size of the model.

In the classification of ArMD disease, the ArConvNet model has achieved 96.88% accuracy, compared to [43] model with 94.03% accuracy, showing an improvement of 2.85%.

## 4. DISCUSSION

The model built in this research uses a layer several times; the repetition of these layers is equal to the spatial dimension of the input. Increasing the number of iterations does not affect the number of parameters and model size, but increases the computational cost and depth of the model. In the experiments conducted by us, increasing the repetition did not increase the accuracy of the model, but in the future, we can look for solutions to prevent this from happening. It is also necessary to conduct tests for the effectiveness of such blocks in 4D inputs such as videos. It seems that the use of the repeated layer has helped to make the model more general, also reduce the parameters, and reduce the fit. Due to the repetition of layers in two spatial dimensions, with the same input data set, this layer receives twice as much variety in the input, which is effective in the model training process.

In the experiments of this research, we sought to evaluate and prove the capabilities of the presented model in identifying a diverse range of features that may be signs of eye disease. However, we can use deep neural networks with general applications in a wide range of activities, including large classification, regression, object recognition, segmentation, and others. In future research, we can evaluate the capabilities of the built model in these fields, including the ImageNet1k [47] dataset. Currently, many hardware companies are trying to implement artificial intelligence programs locally, which improves privacy in medicine [48]. One of the ways to run them locally is to run them in web applications. The smaller size of the model can have a great impact on the faster loading and running of the application.

## 5. CONCLUSION

In this research, we discussed the problem of diagnosing eye diseases and the reasons for their importance. Then, we reviewed the solutions presented in the past and evaluated the research for building deep neural network models for image processing. After that, we presented a new model with the creative idea of iterative use of learning layers. In the following, we explained that the presented method is as if converting the input image into a vector and applying Convolutional layer 1d on it. We continued to show that although at a time, we can process only one image dimension in each layer, the spatial relationships between the image pixels are still preserved. Next, we examined the characteristics of each eye disease. We recognized the diversity of these features in terms of size (small and large signs) dispersion (concentrated and scattered signs) and location with this property that rotation in the feature space did not change the nature of the disease. Then we compared the model with other well-known models and observed the effect of increasing the depth and decreasing the number of model parameters in the results. Finally, we observed that the built model has been able to perform well according to the type of inputs. We evaluated the model on the RfMiD test set and the accuracy was 93.26% in the binary classification section and 67.5% in the odir-2019 disease classification section.

## 6. DECLARATIONS

No funds, grants, or other support was received.

## 7. CONFLICT OF INTEREST

The authors declare no conflict of interest.